\documentclass[letterpaper,10pt,conference]{ieeeconf}

\IEEEoverridecommandlockouts
\usepackage{graphicx}
\usepackage{amsmath}
\usepackage{amssymb}
\usepackage{booktabs}
\usepackage{mathtools}
\usepackage{times}
\usepackage{epsfig}
\usepackage{bbm}
\usepackage{xspace}
\usepackage{float}
\usepackage{placeins}
\usepackage{stfloats}
\usepackage{multicol}
\usepackage{multirow}
\usepackage{makecell}
\usepackage{bm}
\usepackage{etoolbox}
\usepackage{icomma}
\usepackage{array}
\usepackage{tabulary}
\usepackage[table]{xcolor}
\usepackage{paralist}
\usepackage{adjustbox}
\usepackage{caption}
\usepackage{diagbox}

\makeatletter
\newcommand{\setfigurecaptiontype}{\def\@captype{figure}}
\renewcommand{\subsection}{\@startsection{subsection}{2}{\z@}%
  {1.8ex plus 1.3ex minus 0.4ex}%
  {0.85ex plus 0.5ex minus 0ex}%
  {\normalfont\normalsize\itshape}}
\makeatother

\DeclareMathAlphabet\mathbfcal{OMS}{cmsy}{b}{n}

\newcommand{\method}{\textsc{OmniAct3D}\xspace}
\newcommand{\erga}{\textsc{ERGA-Ray}\xspace}
\newcommand{\varc}{\textsc{VARC}\xspace}
\newcommand{\aghe}{\textsc{AGHE}\xspace}

\makeatletter
\let\NAT@parse\undefined
\makeatother

\usepackage[breaklinks,colorlinks]{hyperref}
\hypersetup{
  colorlinks=true,
  citecolor=blue,
  linkcolor=blue,
  urlcolor=blue
}
\usepackage[capitalize]{cleveref}
\crefname{section}{Sec.}{Secs.}
\Crefname{section}{Section}{Sections}
\crefname{table}{Tab.}{Tabs.}
\Crefname{table}{Table}{Tables}

\begin{document}

\title{\LARGE \bf
OmniAct3D: Leveraging Foundation Geometry and Evidence-Grounded Reasoning for Panoramic 3D Detection
}

\author{Runtong Wu$^{1,*}$, Fei Teng$^{1,*}$, Di Wen$^{2}$, Guoqiang Zhao$^{1}$, Kunyu Peng$^{2}$, and Kailun Yang$^{1,\dag}$%
\thanks{This work was supported in part by the National Natural Science Foundation of China (Grant No. 62473139 and No. 62388101), in part by the Hunan Provincial Research and Development Project (Grant No. 2025QK3019), and in part by the State Key Laboratory of Autonomous Intelligent Unmanned Systems (the opening project number ZZKF2025-2-10).}
\thanks{$^{1}$The authors are with the School of Artificial Intelligence and Robotics and the National Engineering Research Center of Robot Visual Perception and Control Technology, Hunan University, China (email: kailun.yang@hnu.edu.cn).}%
\thanks{$^{2}$The authors are with the Institute for Anthropomatics and Robotics, Karlsruhe Institute of Technology, Germany.}
\thanks{$^{*}$Equal contribution.}
\thanks{$^{\dag}$Corresponding author: Kailun Yang.}
}

\maketitle

\begin{abstract}
Accurate 3D detection is essential for mobile embodied agents, while Vision Foundation Models (VFMs) offer transferable visual and geometric priors. Yet existing VFM-based 3D detectors rely on narrow-view monocular images or discrete perspective views, limiting coherent surround perception; equirectangular projection (ERP) instead encodes a continuous $360^\circ$ scene in a single image. Direct transfer remains difficult because ERP organizes geometry and visual information differently, making object-relevant cues hard to model, localize, and preserve. We propose \textbf{\textsc{OmniAct3D}}, a framework that adapts perspective-trained VFM detectors to ERP while preserving transferable VFM priors. To resolve geometric mismatch, the \textbf{ERP-Ray Geometry Adapter (\textsc{ERGA-Ray})} models spherical viewing rays and periodic spatial structure. To localize evidence in scene-wide context, the \textbf{Visual-Action Reasoning Chain (\textsc{VARC})} grounds each hypothesis in relevant panoramic evidence and converts it into a structured geometric action. To recover local cues lost under fixed token budgets, the \textbf{Appearance-Guided Heading Expert (AGHE)} re-encodes object regions at higher resolution for heading estimation. Experiments show that \textsc{OmniAct3D} improves over the previous best 3D detector by $2.96$ NDS points on Spheriverse and over the unadapted VFM baseline by $24.87$ mAP points on PanoMMOcc. With target-specific geometry adaptation, \textsc{VARC} retains $95$--$98\%$ of the same-configuration mAP, indicating reusable object-level 3D reasoning across sensing configurations. The source code will be made publicly available at \url{https://github.com/FeiT-FeiTeng/OmniAct3D}.
\end{abstract}

\section{Introduction}
\label{sec:introduction}
Accurate and robust 3D detection is essential for mobile embodied agents to understand and interact with the physical world.
From autonomous vehicles to mobile robots, agents must estimate the 3D locations, dimensions, and orientations of surrounding objects to support navigation, motion planning, and embodied decision-making~\cite{hu2023planning,sun2025sparsedrive}.

\begin{figure}[!t]
\centering
\includegraphics[width=\columnwidth,pagebox=cropbox]{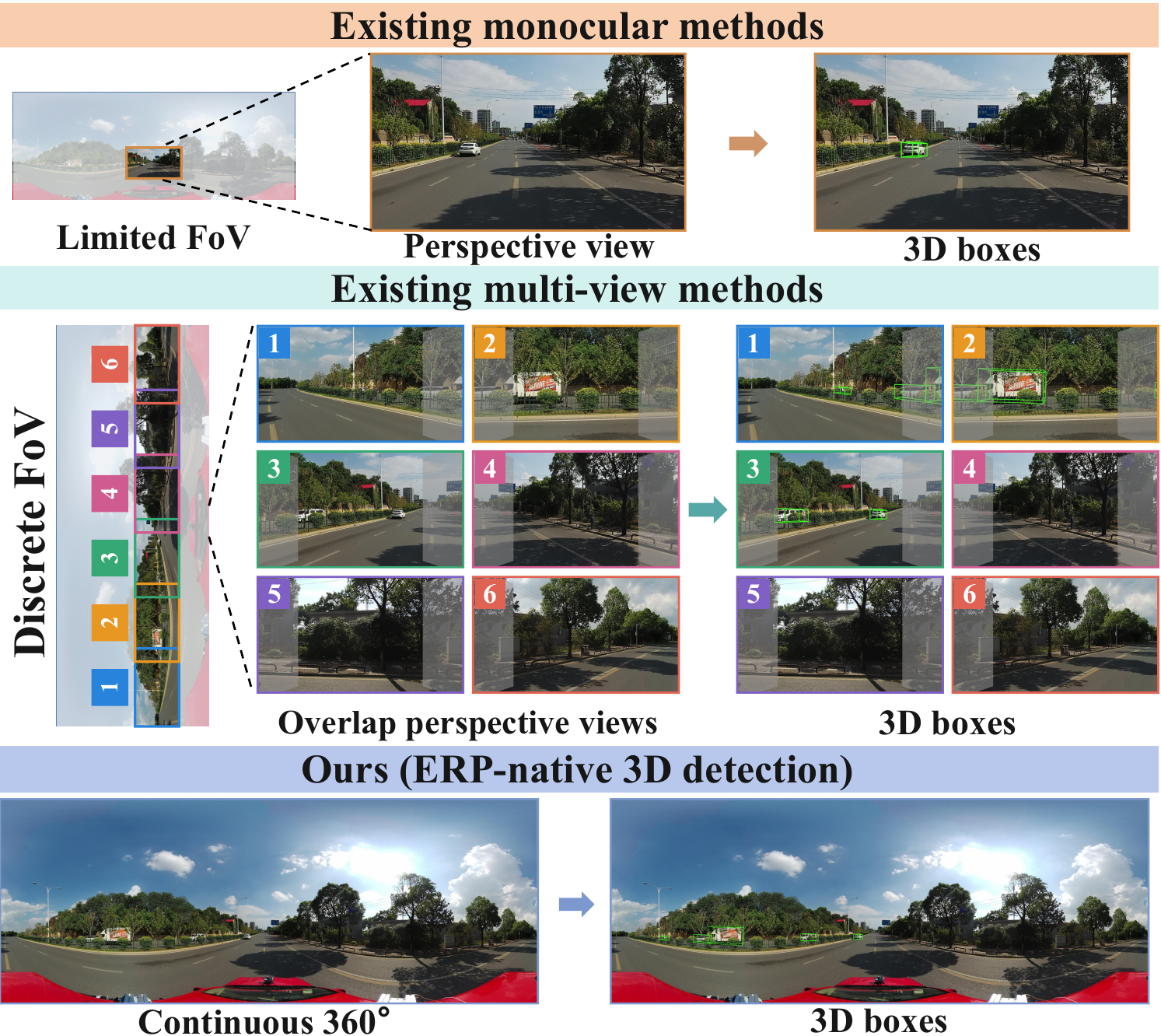}
\caption{Comparison of monocular, multi-view, and ERP 3D detection paradigms. Monocular methods observe a limited field of view~\cite{MonoIA}, while multi-view methods rely on discrete perspective observations~\cite{VGGT}.
In contrast, our method enables coherent $360^\circ$ perception from a single ERP image.
}
\label{fig:paradigm_comparison}
\vskip-2ex
\end{figure}

Recent VFM-based 3D detectors~\cite{fomo3d,MonoDINODETR} leverage transferable visual and geometric priors for object-level 3D perception. Despite this progress, existing methods primarily follow two perspective-image paradigms. As shown in Fig.~\ref{fig:paradigm_comparison}, 
monocular methods~\cite{MonoIA,CHARM3R} infer 3D boxes from a single perspective image but observe only a limited field of view; 
Multi-view methods~\cite{percep,wei2026onebev++} extend the surrounding coverage using discrete perspective observations. 
In contrast, equirectangular projection (ERP) encodes the surrounding spherical field of view within a single image, providing a unified and spatially continuous representation for coherent $360^\circ$ perception~\cite{wu2025quadreamer,teng2026hallucinating}.

However, directly transferring perspective-trained VFM-based 3D detectors to ERP observations remains challenging because ERP organizes geometry and visual information differently from perspective imagery. This representation mismatch makes object-relevant cues difficult to model, localize, and preserve, giving rise to three challenges. First, the geometric priors of VFMs are learned predominantly from locally planar and non-periodic perspective imagery, whereas ERP exhibits spherical viewing directions, latitude-dependent distortion, and horizontal periodicity. Reusing these priors without adaptation therefore produces geometric representations misaligned with ERP structure, particularly at high latitudes and across the horizontal boundary. Second, although a single ERP image captures scene-wide context, the evidence relevant to each object occupies only a small portion of the panorama. Directly decoding 3D hypotheses from global features makes it difficult to localize and associate the supporting evidence for each object, especially when objects are small or occluded. Third, encoding an ultra-wide panorama under a fixed token budget compresses individual objects and discards fine-grained local appearance cues. The loss of features such as front--rear asymmetry impairs directed heading estimation.

To address these challenges, we propose \method, a framework that adapts perspective-trained VFM detectors to ERP observations while preserving transferable VFM priors. To resolve the geometric mismatch, we introduce the ERP-Ray Geometry Adapter (\erga), which models spherical viewing rays and ERP-aware spatial interactions to align pretrained foundation geometry with ERP-specific spherical and periodic structure. To localize object-relevant evidence in scene-wide context, we introduce the Visual-Action Reasoning Chain (\varc), which grounds each 3D hypothesis in relevant panoramic evidence and converts that evidence into a structured geometric action for object-level 3D reasoning. To recover local cues lost under a fixed token budget, we introduce the \aghe, which re-encodes object regions at higher resolution for directed heading estimation.

We evaluate \method on Spheriverse~\cite{CAR} and PanoMMOcc~\cite{DOG}. On Spheriverse, \method achieves $0.1887/0.1825$ mAP/NDS, improving the previous best NDS by $2.96$ points. On PanoMMOcc, it achieves $0.5897/0.4032$ mAP/NDS and improves the unadapted VFM baseline by $24.87$ mAP points. Moreover, with target-specific geometry adaptation via \erga, \varc retains $95\%$ and $98\%$ of same-configuration mAP on Spheriverse and PanoMMOcc, respectively, indicating reusable object-level 3D reasoning across sensing configurations.

In summary, our contributions are as follows:
\begin{itemize}
\setlength{\topsep}{2pt}
\setlength{\partopsep}{0pt}
\setlength{\itemsep}{0pt}
\setlength{\parsep}{0pt}

\item We propose \method, a framework that adapts perspective-trained VFM detectors to ERP observations while preserving transferable VFM priors.

\item We introduce \erga to align foundation geometry with ERP spherical and periodic structure.

\item We introduce \varc to ground 3D hypotheses in relevant panoramic evidence and convert that evidence into a structured geometric action, and \aghe to recover fine-grained local cues for directed heading estimation.

\item Experiments show that \method improves NDS by $2.96$ points over the previous best detector on Spheriverse and mAP by $24.87$ points over the unadapted VFM baseline on PanoMMOcc; with target-specific geometry adaptation, \varc retains $95$--$98\%$ of same-configuration mAP across sensing configurations.
\end{itemize}

\section{Related Work}
\label{sec:related_work}

\subsection{Vision-Based 3D Object Detection}
 
Vision-based 3D object detection estimates object locations, dimensions, and orientations from images. Monocular methods infer 3D states from a single perspective view using geometric constraints, depth estimation, or object-centric representations~\cite{SMOKE,FCOS3D,MonoDLE}. 
Multi-view methods aggregate complementary observations through geometric projection, query-based aggregation, or BEV representations~\cite{DETR3D,BEVFormer,SparseBEV}, while temporal approaches further exploit cross-frame information~\cite{SOLOFusion}.
These methods are developed primarily for perspective imagery. Monocular systems have limited fields of view, whereas surround-view systems rely on multiple discrete camera views and cross-view integration. In contrast, our work studies object-level 3D detection directly from a panoramic representation for continuous $360^\circ$ perception.
 
\subsection{Omnidirectional Vision and ERP Perception}

Equirectangular projection (ERP) represents a spherical field of view within a single image and naturally provides spatially continuous horizontal $360^\circ$ coverage. ERP imagery has been studied for panoramic depth estimation, scene understanding, 3D detection, and omnidirectional occupancy prediction~\cite{OmniDepth,Panoramic3DDetection,OneOcc,O3N}.
However, ERP geometry differs fundamentally from perspective imaging due to latitude-dependent distortion, horizontal periodicity, and spherical viewing rays. Prior work addresses these properties through spherical convolutions and distortion sampling~\cite{SphereNet,SphericalCNN}, projection fusion~\cite{UniFuse}, and panoramic attention~\cite{PanoSwin}. These works highlight the need for ERP-specific geometric modeling for coherent panoramic perception.
 
\subsection{Foundation Models for Panoramic 3D Perception}
Large-scale VFMs provide transferable priors for semantic and geometric perception. Visual models such as DINO~\cite{DINO,DINOv2} and SAM~\cite{SAM,SAM3} learn reusable visual representations, while geometry foundation models such as UniDepth~\cite{UniDepth}, MoGe~\cite{MoGe}, and VGGT~\cite{VGGT} provide transferable priors for depth and 3D structure. We use the term \emph{VFM-based 3D detector} for a 3D detector built on such pretrained models~\cite{MonoDINODETR,OVMono3D,DetAny3D,WildDet3D}. 
Related work also develops camera-general geometry~\cite{DAC,UniK3D,DA2,DAP,UniDAC,PanoVGGT,VGGT-360,OmniVGGT}.
Despite their transferability, adapting VFM-based 3D detection from perspective imagery to ERP remains challenging.
We address this perspective-to-ERP adaptation through ERP geometry alignment and evidence-grounded object-level 3D reasoning.

\newcommand{\mainresultsfloat}{%
\begin{table*}[!t]
\centering
\scriptsize
\setlength{\tabcolsep}{2.3pt}
\renewcommand{\arraystretch}{1.00}
\caption{3D detection results on Spheriverse~\cite{CAR} and PanoMMOcc~\cite{DOG} datasets. Spar., Dens., BEVF., Pola., Solo., DETR., CoiN., PD-B., GeoB., Mono., 3D-M., DetA., and Wild. denote SparseBEV, DenseBEV, BEVFormer, PolarBEVDet, SOLOFusion, DETR3D, CoIn3D, PD-BEV, GeoBEV, MonoCop, 3D-MOOD, DetAny3D, and WildDet3D, respectively.}
\label{tab:car_dog_results}
\begin{adjustbox}{width=\textwidth}
\begin{tabular}{ll*{14}{c}}
\toprule
\textbf{Dataset} & \textbf{Metric}
& \makecell{\textbf{Spar.}\\[-1pt]\cite{SparseBEV}}
& \makecell{\textbf{Dens.}\\[-1pt]\cite{DenseBEV}}
& \makecell{\textbf{BEVF.}\\[-1pt]\cite{BEVFormer}}
& \makecell{\textbf{Pola.}\\[-1pt]\cite{PolarBEVDet}}
& \makecell{\textbf{Solo.}\\[-1pt]\cite{SOLOFusion}}
& \makecell{\textbf{DETR.}\\[-1pt]\cite{DETR3D}}
& \makecell{\textbf{CoiN.}\\[-1pt]\cite{CoIn3D}}
& \makecell{\textbf{PD-B.}\\[-1pt]\cite{GeneralizableBEV}}
& \makecell{\textbf{GeoB.}\\[-1pt]\cite{GeoBEV}}
& \makecell{\textbf{Mono.}\\[-1pt]\cite{MonoCop}}
& \makecell{\textbf{3D-M.}\\[-1pt]\cite{3DMOOD}}
& \makecell{\textbf{DetA.}\\[-1pt]\cite{DetAny3D}}
& \makecell{\textbf{Wild.}\\[-1pt]\cite{WildDet3D}}
& \textbf{\makecell{OmiA.\\(Ours)}}\\
\midrule
\multirow{5}{*}{Spheriverse}
& mAP  $\uparrow$ & \underline{0.1689} & 0.0941 & 0.0941 & 0.1364 & 0.1316 & 0.1296 & 0.1078 & 0.1178 & 0.0654 & 0.0875 & 0.1546 & 0.1183 & 0.0592 & \textbf{0.1887} \\
& mATE $\downarrow$ & 1.0160 & 1.2037 & 1.2051 & \underline{0.9457} & 0.9993 & 1.1622 & 0.9835 & 0.9897 & 1.2248 & 1.1941 & \textbf{0.9087} & 1.1081 & 0.9951 & 0.9677 \\
& mASE $\downarrow$ & 0.6236 & 0.5965 & 0.6645 & 0.6503 & 0.6415 & 0.6524 & 0.6164 & 0.6140 & 0.6117 & 0.3636 & \underline{0.3351} & 0.5726 & 0.3843 & \textbf{0.3052} \\
& mAOE $\downarrow$ & 1.1656 & 1.3051 & 1.2033 & \underline{1.1564} & 1.3356 & 1.3024 & 1.2301 & 1.1955 & 1.2056 & 1.2842 & 1.5658 & 1.6130 & 1.5645 & \textbf{0.8455} \\
& NDS  $\uparrow$ & 0.1221 & 0.0874 & 0.0806 & 0.1086 & 0.1017 & 0.0996 & 0.0939 & 0.0985 & 0.0715 & 0.1074 & \underline{0.1529} & 0.1019 & 0.0917 & \textbf{0.1825} \\
\midrule
\multirow{5}{*}{PanoMMOcc}
& mAP  $\uparrow$ & \underline{0.5758} & 0.5246 & 0.4026 & 0.5520 & 0.5156 & 0.5196 & 0.4713 & 0.4827 & 0.5517 & 0.5504 & 0.3301 & 0.2566 & 0.3410 & \textbf{0.5897} \\
& mATE $\downarrow$ & \underline{0.7092} & 0.8162 & 1.0870 & \textbf{0.7072} & 0.7776 & 0.9311 & 0.8399 & 0.8805 & 0.7252 & 0.7449 & 0.7790 & 0.8358 & 0.9831 & 0.7965 \\
& mASE $\downarrow$ & 0.5996 & 0.6016 & 0.6017 & 0.6111 & 0.6118 & 0.6061 & 0.6078 & 0.6091 & 0.6089 & \textbf{0.2623} & 0.3539 & 0.5967 & 0.3923 & \underline{0.3297} \\
& mAOE $\downarrow$ & 0.9694 & 0.9430 & 0.9576 & 1.2443 & 1.0924 & 0.9085 & \underline{0.8321} & 0.8567 & 0.8852 & 0.9319 & 1.5810 & 1.3586 & 2.1617 & \textbf{0.7902} \\
& NDS  $\uparrow$ & 0.3601 & 0.3262 & 0.2454 & 0.3442 & 0.3189 & 0.3152 & 0.3077 & 0.3067 & 0.3539 & \underline{0.3813} & 0.2518 & 0.1850 & 0.2330 & \textbf{0.4032} \\
\bottomrule
\end{tabular}
\end{adjustbox}
\end{table*}
\vskip-2ex
}

\newcommand{\qualitativeresultsfloat}{%
\begin{figure*}[!t]
\centering
\IfFileExists{Figures/visualization.pdf}{%
  \includegraphics[width=\textwidth,pagebox=cropbox]{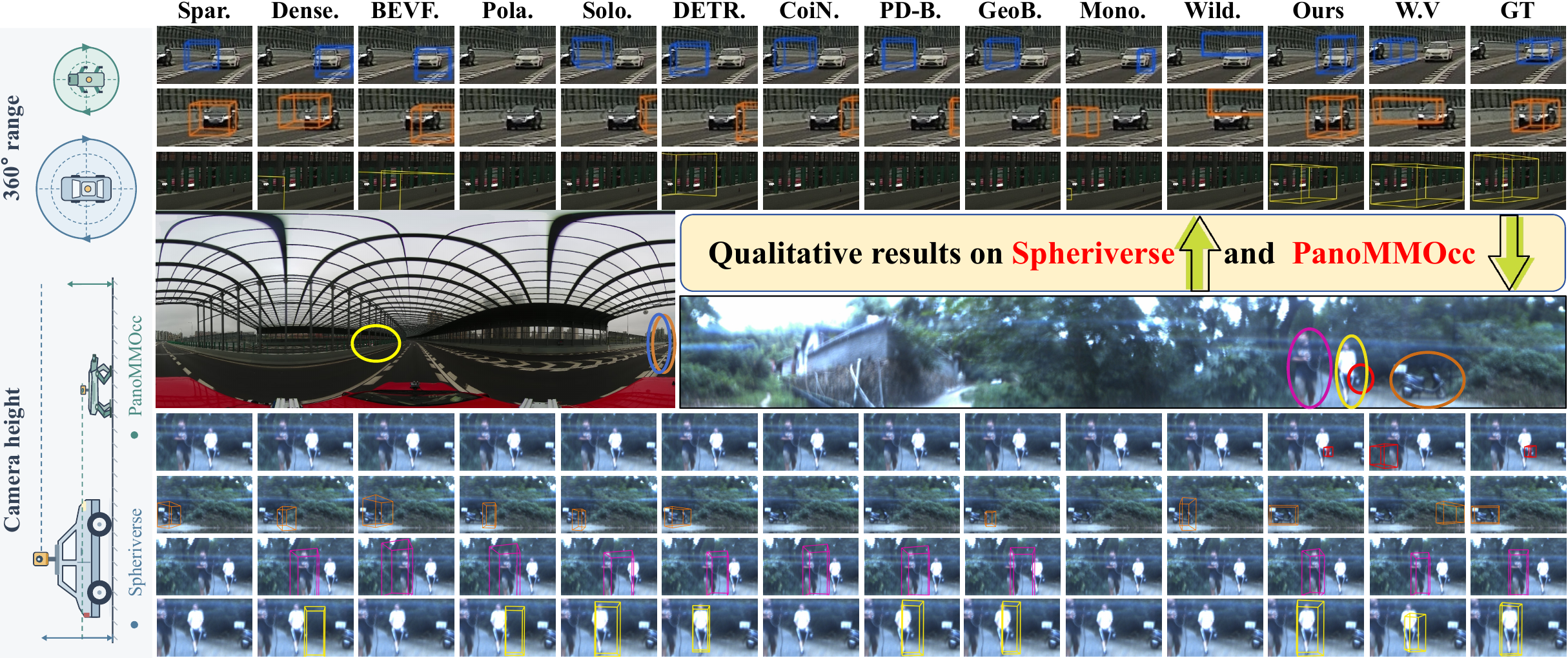}%
}{%
  \fbox{\parbox[c][0.22\textheight][c]{0.96\textwidth}{\centering
  Placeholder for \texttt{Figures/visualization.pdf}}}%
}
%\vskip-1ex
\caption{Qualitative 3D detection comparisons on Spheriverse~\cite{CAR} (top) and PanoMMOcc~\cite{DOG} (bottom). Columns show the Ground Truth (GT), OmniAct3D without VARC (W.V), full OmniAct3D (Ours), and representative baselines.
}
\label{fig:qualitative_results}
\vskip-1ex
\end{figure*}
}

\section{Methodology}
\label{sec:methodology}

Given a full equirectangular projection (ERP) image $I\in\mathbb{R}^{H_I\times W_I\times 3}$, our goal is to directly infer object-level 3D states from a panoramic observation. For each object, the state comprises its 3D center $(X,Y,Z)$, dimensions $(W,L,H)$, orientation $\theta$, category, and detection confidence.
\subsection{Overview of OmniAct3D}

\label{sec:overview}

Given an ERP image $I$, the frozen VFM-based detector extracts two complementary feature representations:
\begin{equation}
\left(F_{\mathrm{vis}},M_0\right)
=
\mathcal{E}_{\mathrm{VFM}}(I),
\qquad
M_0\in\mathbb{R}^{N\times d},
\label{eq:overall_features}
\end{equation}
where $F_{\mathrm{vis}}$ denotes the visual features extracted by the SAM3 branch~\cite{SAM3}, and $M_0$ denotes the pretrained geometry memory produced by the DINOv2-based LingBot branch~\cite{LingBotDepth}. 

\begin{figure*}[t]
\centering
\includegraphics[width=\textwidth,pagebox=cropbox]{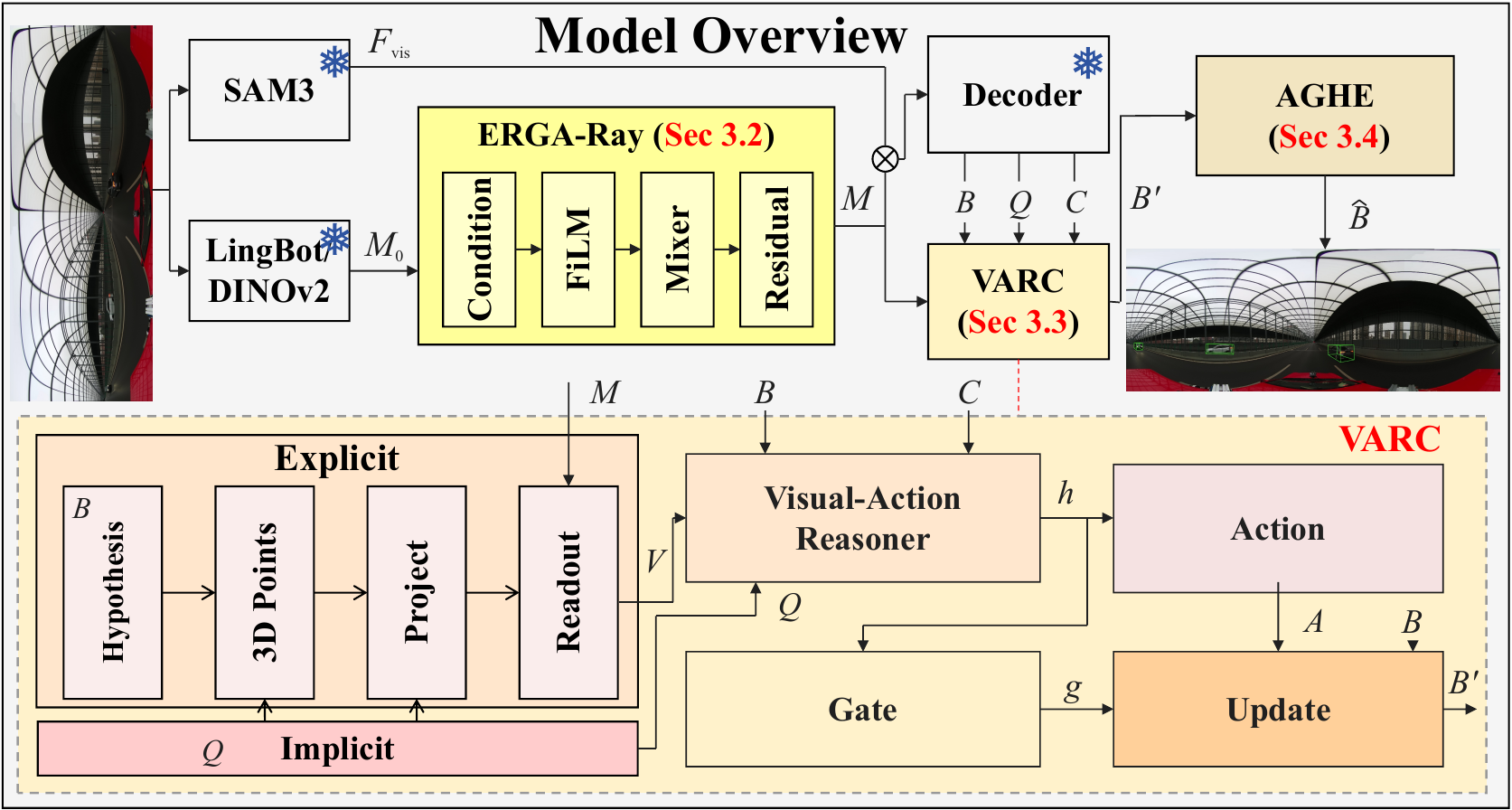}%
\caption{Overview of the feature flow in \method. The frozen VFM-based detector extracts visual features $F_{\mathrm{vis}}$ and pretrained geometry memory $M_0$ from the ERP image. \erga transforms $M_0$ into the ERP-adapted memory $M$, after which the detector produces $B$, $Q$, and $C$. \varc retrieves evidence $V$ and predicts an action $A$ to obtain $B^{\prime}$, while the \aghe updates its directed heading to produce the final prediction $\hat{B}$.}
\label{fig:framework}
\vskip-1ex
\end{figure*}

As illustrated in Fig.~\ref{fig:framework}, the overall pipeline consists of four successive transformations:
\begin{equation}
\begin{array}{@{}l@{}}
M = \mathcal{G}_{\mathrm{ERGA}}
\left(M_0,G_{\mathrm{ERP}}\right),
\quad
(B,Q,C) = \mathcal{D}
\left(F_{\mathrm{vis}},M\right),
\\[2pt]
(V,A,B^{\prime}) = \mathcal{R}_{\mathrm{VARC}}
\left(B,Q,C,M\right),
\quad
\hat{B} = \mathcal{H}_{\mathrm{AGHE}}
\left(I,B^{\prime}\right).
\end{array}
\label{eq:overall_pipeline}
\end{equation}
\noindent\textbf{ERP geometry adaptation.}
$G_{\mathrm{ERP}}$ encodes the spherical viewing rays and periodic positions of the ERP token grid. The \erga operator $\mathcal{G}_{\mathrm{ERGA}}$ uses this encoding to transform the pretrained geometry memory $M_0$ into the detection-oriented ERP geometry memory $M$.

\noindent\textbf{Initial hypothesis generation.}
Conditioned on the visual features $F_{\mathrm{vis}}$ and adapted geometry memory $M$, the detector $\mathcal{D}$ produces an initial 3D box hypothesis $B$, its object query $Q$, and candidate-quality cues
$C=[s_{\mathrm{2D}},s_{\mathrm{3D}},s_{\mathrm{comb}}]$.

\noindent\textbf{Evidence-grounded reasoning.}
Given $(B,Q,C,M)$, \varc retrieves hypothesis-relevant panoramic evidence $V$, reasons jointly over $B$ and $V$ to infer a structured geometric action $A$, and applies $A$ to obtain the resulting box $B^{\prime}$. This explicit \emph{hypothesis--evidence--action} chain grounds each geometric action in relevant panoramic evidence.

\noindent\textbf{Appearance-guided heading estimation.}
The \aghe operator $\mathcal{H}_{\mathrm{AGHE}}$ re-encodes the object region projected onto $I$ at higher resolution and predicts its directed heading from fine-grained local appearance cues, producing the output $\hat{B}$.

The VFM-based detector remains frozen throughout training, whereas \erga, \varc, and \aghe constitute the trainable components for ERP geometry adaptation, evidence-grounded object-level reasoning, and appearance-guided heading estimation, respectively.

\subsection{ERP-Ray Geometry Adapter}

To align pretrained foundation geometry with ERP-specific spherical and periodic structure, we introduce \erga, achieving detection-oriented ERP geometry while preserving transferable priors.

\noindent\textbf{ERP ray geometry.}
Let $(u,v)$ denote normalized ERP coordinates. Longitude and latitude are obtained as $\lambda=\lambda(u)$ and $\phi=\phi(v)$, where the exact vertical mapping depends on the ERP field of view. In the camera coordinate system, spherical ray and lifted 3D points are defined as follows:
\begin{equation}
\resizebox{0.88\columnwidth}{!}{$\displaystyle
\mathbf d(\lambda,\phi)=
\begin{bmatrix}
-\cos\phi\sin\lambda\\
-\sin\phi\\
\cos\phi\cos\lambda
\end{bmatrix},\;
\mathbf P=R\mathbf d,\;
R>0.
$}
\label{eq:erp_ray}
\end{equation}
Here, $R$ denotes the Euclidean viewing-ray range. Conversely, for a known 3D point $\mathbf P$, we obtain $R=\|\mathbf P\|_2$ and $\mathbf d=\mathbf P/R$ in turn. Projection and back-projection wrap across the ERP seam.

\noindent\textbf{Ray-conditioned panoramic adaptation.}
Foundation geometry does not explicitly encode ERP-specific viewing directions, latitude-dependent distortion, or horizontal periodicity. \erga therefore associates each token $M_{0,i}$ with:
\begin{equation}
\mathbf g_i=
\big[
d_x,d_y,d_z,\,
\sin\lambda,\cos\lambda,\,
\sin\phi,\cos\phi
\big].
\label{eq:erp_condition}
\end{equation}
The ray components encode the physical viewing direction, while the trigonometric angular terms provide a smooth spherical positional encoding.
In particular, $\cos\phi$ also reflects the latitude-dependent area distortion of the ERP parameterization and therefore serves as a distortion cue.

A lightweight geometry encoder maps $\mathbf g_i$ to $\mathbf e_i$, which modulates a low-dimensional projection of the foundation token through FiLM:
\begin{equation}
\resizebox{0.88\columnwidth}{!}{$\displaystyle
\begin{gathered}
\tilde{\mathbf c}_i
=\left(1+\eta\tanh\boldsymbol\gamma_i\right)
\odot W_d\operatorname{LN}(M_{0,i})
+\boldsymbol\beta_i+\mathbf e_i,\\
(\boldsymbol\gamma_i,\boldsymbol\beta_i)
=f_{\mathrm{film}}(\mathbf e_i).
\end{gathered}
$}
\label{eq:ray_film}
\end{equation}

The ray-conditioned tokens are rearranged into their ERP feature layout and processed by an asymmetric spatial mixer. Horizontal interactions use circular padding to preserve left--right continuity, whereas vertical interactions remain non-periodic. 
\erga predicts a gated residual over the pretrained geometry memory:
\begin{equation}
\begin{aligned}
M={}&M_0+\mathbf g_m\odot W_u
\operatorname{Mix}_{\mathrm{ERP}}(\tilde{\mathbf C}),\\
\mathbf g_m={}&
\sigma\!\left(
f_{\mathrm{gate}}(\tilde{\mathbf C})
\right).
\end{aligned}
\label{eq:erga_residual}
\end{equation}
Here, $W_d$ and $W_u$ are learned down- and up-projection matrices,$\eta$ bounds the multiplicative FiLM modulation, and
$\tilde{\mathbf C}=[\tilde{\mathbf c}_1,\ldots,\tilde{\mathbf c}_N]$denotes the stacked ray-conditioned tokens.
This residual formulation preserves transferable foundation priors while introducing the spherical and periodic geometry required by panoramic 3D object detection.

\subsection{Visual-Action Reasoning Chain}

To ground 3D hypotheses in panoramic evidence, we introduce \varc, enabling evidence-grounded object-level 3D reasoning through structured geometric actions.

\noindent\textbf{Hypothesis representation.}
\varc takes an explicit box hypothesis $B$ and its object query $Q$. 
We parameterize $B$ in panoramic coordinates:
\begin{equation}
\begin{aligned}
s={}&[\log\rho,\lambda,\phi,\log W,
\log L,\log H,\theta_{\mathrm{obs}}],\\
\rho={}&\sqrt{X^2+Z^2},\qquad
\theta_{\mathrm{obs}}=\operatorname{wrap}(\theta-\lambda).
\end{aligned}
\label{eq:spherical_state}
\end{equation}
with
$\lambda=\operatorname{atan2}(-X,Z)$ and
$\phi=\operatorname{atan2}(-Y,\rho)$.
Here, $\rho$ is the horizontal range used by \varc, while $R$ is the Euclidean viewing-ray range used by ERP lifting and \erga supervision; they satisfy $\rho=R\cos\phi$. We use trigonometric embeddings $\psi(s)$ for angular variables before feeding the state to the Reasoner.
The box represents the geometric hypothesis, while $Q$ supplies object-level context.

\noindent\textbf{Hypothesis-grounded panoramic evidence.}
The evidence reader grounds each 3D hypothesis in target-relevant panoramic observations. 
For each $B$, we construct $K=13$ object-centric support points comprising the box center, eight corners, and four side-face centers.
Query-conditioned offsets adapt this geometric template while remaining bounded, \textit{i.e.}, $\|\Delta P\|_\infty\leq\delta_{\max}$. 
The resulting points are projected to ERP coordinates and sampled from the fixed geometry memory using circular interpolation. 
For the $k$-th support point:
\begin{equation}
\begin{aligned}
m^k={}&
\mathcal{S}_{\mathrm{circ}}\!\left(
M,\Pi_{\mathrm{ERP}}(P^k;B)
\right),\\
a^k={}&
\operatorname{softmax}_{k}\!\left(
\frac{f_k(m^k)^\top f_q(Q)}{\sqrt d}
\right).
\end{aligned}
\label{eq:evidence_sampling}
\end{equation}
We further define the projection descriptor
$p=[w^{\mathrm{ang}},h^{\mathrm{ang}},s^{\mathrm{seam}},r^{\mathrm{vis}}]$, which encodes angular width, angular height, seam crossing, and the fraction of visible support points. 
The target-relevant panoramic evidence is then represented as:
\begin{equation}
V=
f_{\mathrm{read}}\!\left(
\left[
\begin{aligned}
&\sum_{k=1}^{K}a^k f_v(m^k),\,
f_q(Q),\\
&f_s(\psi(s)),\,f_p(p)
\end{aligned}
\right]
\right).
\label{eq:evidence_reading}
\end{equation}
Thus, $V$ provides hypothesis-grounded object-level evidence from the panoramic representation.

\noindent\textbf{Evidence-driven geometric action.}
Given $V$, the \emph{Visual-Action Reasoner} combines the evidence with object query, spherical state, quality cues, and projection descriptor:
\begin{equation}
h =
f_{\mathrm{reason}}\!\left(
\left[
\begin{aligned}
&V,\,Q,\,f_s(\psi(s)),\\
&f_c(C),\,f_p(p)
\end{aligned}
\right]
\right).
\label{eq:reasoner}
\end{equation}
From $h$, the model predicts a structured action $A$ and gate
$g=\sigma(f_{\mathrm{gate}}(h))$, which controls the action strength.
We factorize the horizontal-range action into a discrete direction $d\in\{-1,0,+1\}$ and continuous magnitude $\mu\in[0,1]$:
\begin{equation}
\Delta\log\rho
=g\,d\,\mu\,\log(1.15).
\label{eq:radial_action}
\end{equation}
Here, ``radial'' refers to the horizontal range $\rho$, rather than the Euclidean viewing-ray range $R$ used for ERP lifting. During training, $d$ is represented by the expectation under the three-class distribution, whereas inference uses argmax.

Longitude, latitude, dimensions, and observation-relative yaw use bounded actions:
\begin{equation}
A^{\mathrm{cont}}
=
g\,
\tanh
\left(
f_{\mathrm{act}}(h)
\right)
\odot
\boldsymbol{\kappa},
\label{eq:continuous_action}
\end{equation}
where $\boldsymbol{\kappa}$ provides variable-dependent geometric bounds.
These terms form $A$, yielding:
\begin{equation}
s^{\prime}=\operatorname{WrapClamp}(s+A).
\label{eq:geometric_action}
\end{equation}
Here, $\operatorname{WrapClamp}$ wraps longitude and observation-relative yaw to their periodic intervals, clamps latitude to the calibrated vertical field of view, and preserves the log-parameterization of range and dimensions.
Inverse parameterization yields the resulting 3D box $B^{\prime}$.

\mainresultsfloat

\subsection{Appearance-Guided Heading Expert}
\label{sec:heading_expert}

To recover fine-grained heading cues weakened by global ERP encoding, we introduce the \aghe for directed heading estimation via high-resolution object-centric re-encoding.

The \aghe operates on the resulting 3D boxes.
Rather than rerunning the complete VFM-based 3D detector, it re-encodes only the projected object regions at a higher object-centric resolution using the pretrained DINOv2 encoder~\cite{DINOv2} from the LingBot branch. 
For each box $B_i^{\prime}$, we project it back to the original ERP image, crop the corresponding object region with contextual margins, and reuse the pretrained DINOv2 encoder~\cite{DINOv2}:
\begin{equation}
\begin{gathered}
I_i^{\mathrm{crop}}
=\operatorname{Crop}\!\left(
I,\Pi_{\mathrm{ERP}}(B_i^{\prime})
\right),\\
F_i^{\mathrm{local}}
=\mathcal{F}_{\mathrm{DINO}}
(I_i^{\mathrm{crop}}).
\end{gathered}
\label{eq:local_heading_feature}
\end{equation}
All crops from each ERP image are batched; SAM3~\cite{SAM3}, the depth decoder, and the detection head are not re-executed.

The local appearance is combined with a category embedding and an orientation-free geometric encoding:
\begin{equation}
\mathbf g_i^{\mathrm{head}}
=
\left[
\begin{aligned}
&\log\rho_i,\,\sin\lambda_i,\cos\lambda_i,\\
&\log W_i,\log L_i,\log H_i
\end{aligned}
\right].
\label{eq:heading_geometry}
\end{equation}
The \aghe predicts periodic observation-relative heading and converts it to the camera-frame direction:
\begin{equation}
\begin{aligned}
(\hat s_i,\hat c_i)
&=f_{\mathrm{head}}\!\left(
F_i^{\mathrm{local}},\mathbf e_i^{\mathrm{cls}},
\mathbf g_i^{\mathrm{head}}
\right),\\
\hat\theta_i
&=\operatorname{wrap}\!\left(
\operatorname{atan2}(\hat s_i,\hat c_i)+\lambda_i
\right).
\end{aligned}
\label{eq:heading_prediction}
\end{equation}
The expert replaces the orientation while preserving the center, dimensions, category, and detection score.

\subsection{Training Objectives}
\erga is supervised by a multi-target loss:
\begin{equation}
\begin{aligned}
\mathcal L_{\mathrm{ERGA}}
={}&\mathcal L_{\mathrm{target}}
+\lambda_{\mathrm{ori}}\mathcal L_{\mathrm{ori}}
+\lambda_{\mathrm{pre}}\mathcal L_{\mathrm{pre}}\\
&+\lambda_{\mathrm{conf}}\mathcal L_{\mathrm{conf}}
+\lambda_{\mathrm{res}}\mathcal L_{\mathrm{res}}.
\end{aligned}
\label{eq:erga_loss}
\end{equation}
Here, $\mathcal L_{\mathrm{target}}$ jointly regresses the seam-aware image-center residual, log-range $\log R$, and log-dimensions. 
The remaining terms preserve orientation, pretrained regression, and confidence predictions, while constraining the magnitude of the learned adaptation residual.
\varc is supervised by the following objective:
\begin{equation}
\begin{aligned}
\mathcal L_{\mathrm{VARC}}
={}&\mathcal L_{b}
+\lambda_{\mathrm{id}}\mathcal L_{\mathrm{id}}
+\lambda_{g}\mathcal L_{g}
+\lambda_{r}\mathcal L_{r}.
\end{aligned}
\label{eq:varc_loss}
\end{equation}
Here, $\mathcal L_b$ supervises the resulting spherical box state, $\mathcal L_{\mathrm{id}}$ encourages identity actions for already accurate hypotheses, and $\mathcal L_g$ supervises action-gate activation.
$\mathcal L_r$ denotes the radial loss, combining direction classification, action ranking, and action-magnitude regression.
The \aghe is supervised on the sine-cosine encoding of the periodic observation-relative heading in Eq.~\eqref{eq:heading_prediction} using a cosine loss.

\section{Experiments}

We evaluate \method on two panoramic 3D detection benchmarks. We then analyze the effects of \erga, \varc, and \aghe, and test whether \varc transfers between different sensing configurations.

\subsection{Experimental Setup}

\noindent\textbf{Datasets and metrics.}
We use Spheriverse~\cite{CAR} and PanoMMOcc~\cite{DOG}, which were originally developed for 3D occupancy prediction. Both datasets provide LiDAR observations, 3D bounding boxes, and camera calibration. We transform the boxes from the LiDAR coordinate system to the panoramic camera coordinate system to construct the two ERP 3D detection benchmarks. The datasets differ in camera height, vertical field of view, and object-distance distribution. Following the nuScenes-style protocol~\cite{nuScenes}, we report mAP, NDS, mATE, mASE, and mAOE as evaluation metrics. The two benchmarks do not provide the attribute and velocity annotations required for mAAE and mAVE. We therefore report a three-TP variant of NDS.

\noindent\textbf{Baselines and evaluation protocol.}
We compare against representative 3D detectors spanning Cartesian-grid BEV architectures~\cite{BEVFormer,SOLOFusion,DenseBEV}, geometry-oriented BEV architectures~\cite{GeneralizableBEV,GeoBEV}, polar-coordinate BEV architectures~\cite{PolarBEVDet}, sparse query-based architectures~\cite{DETR3D,SparseBEV}, configuration-aware architectures~\cite{CoIn3D}, monocular detectors~\cite{MonoCop}, and VFM-based detectors~\cite{DetAny3D,WildDet3D,3DMOOD}.
In Table~\ref{tab:car_dog_results}, WildDet3D uses its original square preprocessing scheme, while 3D-MOOD and DetAny3D decompose each panorama into perspective views.
The remaining methods operate directly on the full rectangular equirectangular projection (ERP) panorama.
All ablations use the same ERP protocol. Unless otherwise noted, all methods are evaluated on both benchmarks with only minimal dataset-specific adaptations.

\noindent\textbf{Implementation details.}
All experiments are conducted on four NVIDIA GeForce RTX 3090 GPUs. The VFM-based 3D detector remains frozen throughout training, and all input images are resized to $2480 \times 512$ pixels. We optimize \erga, \varc, and \aghe separately, updating one component at a time while keeping the others fixed. \erga and \varc together introduce approximately $17.07$M trainable parameters, corresponding to only a $0.15\%$ increase in the total parameter count, while incurring a measured inference-latency overhead of $3.3\%$--$5.0\%$.
For VARC transfer, we use a target-specific ERGA-Ray module. 

\subsection{Main Results}

To evaluate the overall detection performance of \method, we compare it with existing 3D detectors on both benchmarks. Table~\ref{tab:car_dog_results} reports the results. On Spheriverse, \method improves the previous best mAP and NDS by $1.98$ and $2.96$ points, respectively ($0.1887$ vs.\ $0.1689$ and $0.1825$ vs.\ $0.1529$). On PanoMMOcc, it improves the previous best mAP and NDS by $1.39$ and $2.19$ points, respectively ($0.5897$ vs.\ $0.5758$ and $0.4032$ vs.\ $0.3813$). It also gives the lowest mASE and mAOE on Spheriverse. The consistent gains across both benchmarks show the effectiveness of \method under different panoramic sensing configurations.
Fig.~\ref{fig:qualitative_results} qualitatively compares OmniAct3D without VARC (W.V), full OmniAct3D (Ours), and representative baselines across diverse Spheriverse and PanoMMOcc scenes.

\qualitativeresultsfloat

\begin{table}[ht]
\centering
\scriptsize
\setlength{\tabcolsep}{2.6pt}
\renewcommand{\arraystretch}{1.00}
\caption{Component ablation.}
\label{tab:main_ablation}
%\vskip-1ex
\begin{adjustbox}{max width=\columnwidth}
\begin{tabular}{lccc*{5}{c}}
\toprule
\textbf{Dataset}
& \textbf{ERGA}
& \textbf{VARC}
& \makecell{\textbf{AGHE}}
& \textbf{mAP} $\uparrow$
& \textbf{NDS} $\uparrow$
& \textbf{mATE} $\downarrow$
& \textbf{mASE} $\downarrow$
& \textbf{mAOE} $\downarrow$ \\
\midrule

\multirow{5}{*}{Spheriverse}
& $\times$ & $\times$ & $\times$
& 0.0301 & 0.0335 & 1.1216 & 0.8156 & \underline{1.1919} \\

& $\checkmark$ & $\times$ & $\times$
& \underline{0.1191} & 0.1168 & \underline{1.0127} & 0.4272 & 1.3631 \\

& $\times$ & $\checkmark$ & $\times$
& 0.0798 & 0.1073 & 1.2703 & \underline{0.3257} & 1.2984 \\

& $\checkmark$ & $\checkmark$ & $\times$
& \textbf{0.1887} & \underline{0.1670} & \textbf{0.9677}
& \textbf{0.3052} & 1.5585 \\

& $\checkmark$ & $\checkmark$ & $\checkmark$
& \textbf{0.1887} & \textbf{0.1825} & \textbf{0.9677}
& \textbf{0.3052} & \textbf{0.8455} \\

\midrule

\multirow{5}{*}{PanoMMOcc}
& $\times$ & $\times$ & $\times$
& 0.3410 & 0.2330 & 0.9831 & 0.3923 & 1.0770 \\

& $\checkmark$ & $\times$ & $\times$
& \underline{0.4353} & 0.2861 & 0.9707 & 0.3445 & 2.1814 \\

& $\times$ & $\checkmark$ & $\times$
& 0.2920 & 0.2442 & \underline{0.8779} & \underline{0.3392} & \underline{0.8009} \\

& $\checkmark$ & $\checkmark$ & $\times$
& \textbf{0.5897} & \underline{0.3822} & \textbf{0.7965}
& \textbf{0.3297} & 2.0781 \\

& $\checkmark$ & $\checkmark$ & $\checkmark$
& \textbf{0.5897} & \textbf{0.4032} & \textbf{0.7965}
& \textbf{0.3297} & \textbf{0.7902} \\

\bottomrule
\end{tabular}
\end{adjustbox}
\end{table}

\section{Analysis and Ablation Studies}
We analyze equirectangular projection (ERP) geometry adaptation, hypothesis-evidence-action reasoning, fine-grained heading recovery, and reasoning reuse.

\subsection{Component Ablation Analysis}
\label{sec:component_analysis}
Table~\ref{tab:main_ablation} reports the component ablation. Adding \varc to \erga improves mAP by $6.96$ points on Spheriverse ($0.1887$ vs.\ $0.1191$) and by $15.44$ points on PanoMMOcc ($0.5897$ vs.\ $0.4353$). Adding \aghe further reduces mAOE while leaving mAP unchanged. Thus, \varc mainly improves object localization, whereas \aghe improves heading estimation in full \method.

\subsection{ERP Geometry Adaptation}
\label{sec:erga_analysis}

\noindent\textbf{ERP-specific geometry modeling.}
To investigate whether existing adaptation methods are sufficient for ERP 3D detection, Table~\ref{tab:erga_ablation} compares several methods using the same LingBot backbone~\cite{LingBotDepth}. No Adapt.\ changes only the projection to ERP and directly uses the original LingBot features. MLP and AdaptFormer~\cite{AdaptFormer} apply feature adaptation, while Circular Conv~\cite{CircularCNN}, Ray-FiLM~\cite{FiLM}, and the PanoSwin-based~\cite{PanoSwin} model horizontal periodicity, spherical viewing rays, and panoramic spatial interactions, respectively. \erga combines spherical ray conditioning with periodic spatial modeling. Overall, \erga improves mAP/NDS over the strongest alternatives by $21.5\%/9.1\%$ on Spheriverse and by $16.7\%/11.1\%$ on PanoMMOcc.

\noindent\textbf{Geometry backbone comparison.}
To test whether \erga is effective with different geometry backbones, Table~\ref{tab:erga_ablation} also evaluates UniK3D~\cite{UniK3D}, DA$^2$~\cite{DA2}, and DAP~\cite{DAP}. UniK3D is designed for general camera models, while DA$^2$ and DAP are designed for panoramic images. Although DAP with \erga has lower mAP than LingBot with \erga, adding \erga still improves DAP by $44.4\%$ on Spheriverse ($0.0986$ vs.\ $0.0683$) and by $13.1\%$ on PanoMMOcc ($0.2097$ vs.\ $0.1854$). This shows that \erga is also effective with a different backbone.

\subsection{Hypothesis-Evidence-Action Reasoning}
\label{sec:varc_analysis}

\begin{figure*}[!t]
\centering
\includegraphics[width=\textwidth,pagebox=cropbox]{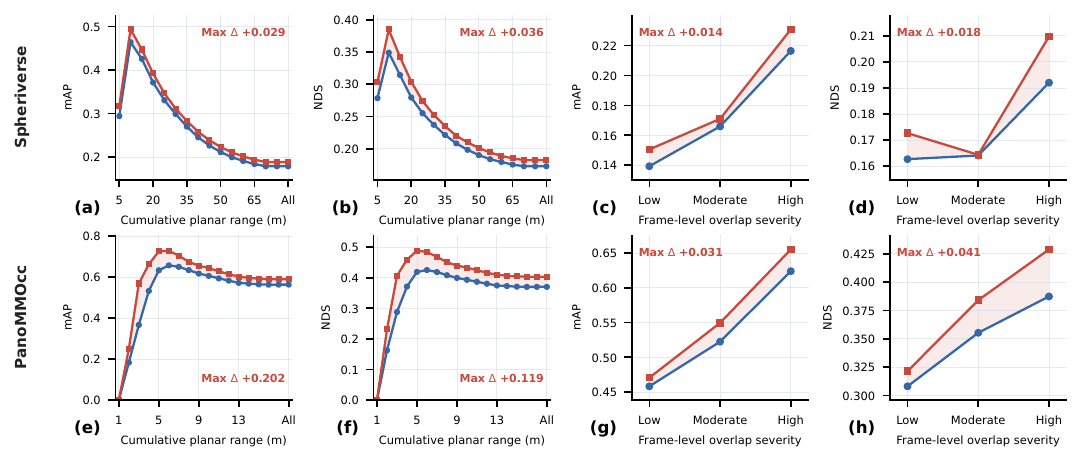}
\vskip-1.5ex
\caption{
Cumulative mAP and NDS are reported for Spheriverse~\cite{CAR} (a--d) and PanoMMOcc~\cite{DOG} (e--h) across planar-range thresholds (a,b,e,f) and overlap levels (c,d,g,h). 
Blue curves denote the ERGA-Ray+AGHE variant without VARC, whereas red curves denote full OmniAct3D.
}
\label{fig:distance}
% \vskip-1ex
\end{figure*}

To examine how \varc improves object-level predictions, we analyze difficult targets, its three main components, and the resulting localization changes.

\begin{table}[ht]
\centering
\scriptsize
\setlength{\tabcolsep}{2.6pt}
\renewcommand{\arraystretch}{1.00}
\caption{Geometry adaptation and backbone comparisons.
}
%\vskip-1ex
\label{tab:erga_ablation}
\begin{tabular*}{\columnwidth}{@{\extracolsep{\fill}}llcc@{}}
\toprule
\textbf{Method} & \textbf{Backbone}
& \shortstack{\textbf{Spheriverse}\\\textbf{mAP / NDS}}
& \shortstack{\textbf{PanoMMOcc}\\\textbf{mAP / NDS}} \\
\midrule
No Adapt. & LingBot~\cite{LingBotDepth} & 0.0301 / 0.0335 & 0.3410 / 0.2330 \\
MLP & LingBot~\cite{LingBotDepth} & \underline{0.0980} / 0.1066 & \underline{0.3730} / \underline{0.2575} \\
AdaptFormer~\cite{AdaptFormer} & LingBot~\cite{LingBotDepth} & 0.0959 / 0.1060 & 0.3585 / 0.2458 \\
Circular Conv~\cite{CircularCNN} & LingBot~\cite{LingBotDepth} & 0.0975 / \underline{0.1071} & 0.3692 / 0.2522 \\
Ray-FiLM~\cite{FiLM} & LingBot~\cite{LingBotDepth} & 0.0929 / 0.1051 & 0.3689 / 0.2534 \\
PanoSwin-based~\cite{PanoSwin} & LingBot~\cite{LingBotDepth} & 0.0948 / 0.1056 & 0.3690 / 0.2552 \\
\textbf{ERGA-Ray} & LingBot~\cite{LingBotDepth}
& \textbf{0.1191 / 0.1168} & \textbf{0.4353 / 0.2861} \\
\midrule
UniK3D Direct & UniK3D~\cite{UniK3D} & 0.0682 / \underline{0.0948} & 0.1556 / 0.0778 \\
DA$^2$ Direct & DA$^2$~\cite{DA2} & 0.0302 / 0.0518
& \textbf{0.2621 / 0.1919} \\
DAP Direct & DAP~\cite{DAP} & \underline{0.0683} / 0.0947 & 0.1854 / 0.1315 \\
\textbf{DAP + ERGA} & DAP~\cite{DAP}
& \textbf{0.0986 / 0.1174} & \underline{0.2097} / \underline{0.1696} \\
\bottomrule
\end{tabular*}
\vskip-1ex
\end{table}

\noindent\textbf{Evidence-challenging targets.}
To test \varc on evidence-challenging targets, we use a subset containing objects in the lowest area quartile that are at least $75\%$ occluded by a closer object. The complement contains all remaining objects, and the main results use the full validation sets. Center@8 is the one-to-one center error capped at $8\,\mathrm{m}$, while S@2m and S@4m measure localization success at the corresponding thresholds. Diagnostic mAP is computed on this subset. Table~\ref{tab:evidence_challenging_subset} compares the frozen detector with \erga only on the subset and its complement. The subset has higher center error and lower S@2m on both datasets. The reported differences are the subset values minus the complement values, with all $95\%$ frame-cluster bootstrap confidence intervals excluding zero.

\begin{table}[ht]
\centering
\scriptsize
\setlength{\tabcolsep}{2.6pt}
\renewcommand{\arraystretch}{1.00}
\caption{Challenging targets versus complement.}
\label{tab:evidence_challenging_subset}
%\vskip-1ex
\begin{adjustbox}{max width=\columnwidth}
\begin{tabular}{llccc}
\toprule
\textbf{Dataset} & \textbf{Metric}
& \textbf{Complement} & \textbf{Evidence-chall.}
& \textbf{Difference [95\% CI]} \\
\midrule
\multirow{2}{*}{Spheriverse}
& Center@8 $\downarrow$ & 5.298\,m & 6.924\,m
& $+1.626$ $[1.396,1.847]$\,m \\
& S@2m $\uparrow$ & 23.1\% & 3.7\%
& $-19.5$ $[-21.5,-17.3]$ pp \\
\midrule
\multirow{2}{*}{PanoMMOcc}
& Center@8 $\downarrow$ & 0.957\,m & 1.942\,m
& $+0.985$ $[0.799,1.179]$\,m \\
& S@2m $\uparrow$ & 88.7\% & 59.0\%
& $-29.7$ $[-35.8,-23.8]$ pp \\
\bottomrule
\end{tabular}
\end{adjustbox}
\vskip-1ex
\end{table}

\noindent\textbf{Reasoning-chain components.}
Table~\ref{tab:varc_design_ablation} ablates the key VARC designs by replacing hypothesis-conditioned evidence with fixed evidence, structured geometric actions with unstructured regression, and object-query context with box-state-only conditioning.
Full \varc achieves $0.1768$ diagnostic mAP on PanoMMOcc. Removing evidence localization, structured action, or object-query context reduces it to $0.1257$, $0.1400$, and $0.1311$, respectively. All three components therefore contribute to the final result.

\begin{table}[ht]
\centering
\scriptsize
\setlength{\tabcolsep}{2.8pt}
\renewcommand{\arraystretch}{1.00}
\caption{VARC component ablation.}
\label{tab:varc_design_ablation}
%\vskip-1ex
\begin{tabular*}{\columnwidth}{@{\extracolsep{\fill}}lcc|ccc@{}}
\toprule
& \multicolumn{2}{c|}{\textbf{Spheriverse}}
& \multicolumn{3}{c}{\textbf{PanoMMOcc}} \\
\cmidrule(lr){2-3}\cmidrule(lr){4-6}
\textbf{Variant}
& mAP $\uparrow$
& NDS $\uparrow$
& mAP $\uparrow$
& NDS $\uparrow$
& Diag. mAP $\uparrow$ \\
\midrule

Fixed Evidence
& \underline{0.1199} & \underline{0.1274}
& \underline{0.5892} & \underline{0.4034} & 0.1257 \\

Unstruct. Action
& 0.1180 & 0.1266
& 0.5890 & \textbf{0.4037} & \underline{0.1400} \\

BBox State Only
& 0.1194 & 0.1271
& 0.5889 & 0.4015 & 0.1311 \\

\textbf{Full VARC}
& \textbf{0.1887}
& \textbf{0.1825}
& \textbf{0.5897}
& 0.4032
& \textbf{0.1768} \\

\bottomrule
\end{tabular*}
\vskip-2ex
\end{table}

\noindent\textbf{Geometric gains on evidence-challenging targets.}
To measure the localization changes produced by \varc, we compare the same targets with and without \varc. Center Gain is the reduction in center error, $\Delta$S is the change in localization success, and positive-change precision is the percentage of changed predictions that improve. As shown in Table~\ref{tab:varc_geometric_gains}, \varc reduces center error by $0.80\,\mathrm{m}$ on Spheriverse and $0.54\,\mathrm{m}$ on PanoMMOcc, with positive-change precision above $86\%$.

\begin{table}[ht]
\centering
\scriptsize
\setlength{\tabcolsep}{3.2pt}
\renewcommand{\arraystretch}{1.00}
\caption{VARC gains on evidence-challenging targets.}
\label{tab:varc_geometric_gains}
%\vskip-1ex
\begin{tabular*}{\columnwidth}{@{\extracolsep{\fill}}lcccc@{}}
\toprule
\textbf{Dataset}
& \makecell{Center\\Gain $\uparrow$}
& $\Delta$S@2m $\uparrow$
& $\Delta$S@4m $\uparrow$
& \makecell{Positive-change\\Prec. $\uparrow$} \\
\midrule
Spheriverse
& 0.8019 m
& +3.79\%
& +12.30\%
& 86.79\% \\

PanoMMOcc
& 0.5430 m
& +13.97\%
& +7.94\%
& 87.88\% \\
\bottomrule
\end{tabular*}
\vskip-2ex
\end{table}

We further examine its effectiveness across different object ranges and scene-overlap levels. 
Figure~\ref{fig:distance} shows that adding \varc improves mAP and NDS across different object ranges and overlap levels. The gains are larger for nearby objects and scenes with more overlap.

\subsection{Fine-Grained Heading Recovery}
We evaluate whether object-centric re-encoding recovers local appearance cues weakened by global ERP encoding. 
Table~\ref{tab:main_ablation} shows that the \aghe leaves mAP unchanged while reducing mAOE from $1.5585$ to $0.8455$ on Spheriverse and from $2.0781$ to $0.7902$ on PanoMMOcc, increasing NDS to $0.1825$ and $0.4032$. 
Thus, \aghe improves heading estimation without changing object localization.

\begin{table}[H]
\centering
\scriptsize
\setlength{\tabcolsep}{3.0pt}
\renewcommand{\arraystretch}{1.08}
\caption{Cross-configuration transfer of VARC.}
\label{tab:cross_platform_transfer}
%\vskip-1ex
\begin{tabular*}{\columnwidth}{@{\extracolsep{\fill}}lcccc@{}}
\toprule
& \multicolumn{2}{c}{\textbf{Spheriverse Eval}}
& \multicolumn{2}{c}{\textbf{PanoMMOcc Eval}} \\
\cmidrule(lr){2-3}\cmidrule(lr){4-5}
\textbf{VARC Train}
& mAP $\uparrow$ & NDS $\uparrow$
& mAP $\uparrow$ & NDS $\uparrow$ \\
\midrule
\textbf{Spheriverse} & 0.1887 & 0.1825 & 0.5803 & 0.3833 \\
\textbf{PanoMMOcc} & 0.1800 & 0.1727 & 0.5897 & 0.4032 \\
\bottomrule
\end{tabular*}
\vskip-2ex
\end{table}

\subsection{Cross-Configuration Reasoning Transfer}
To investigate whether \varc remains reusable across sensing configurations, we transfer it between the two datasets without updating its parameters, using the target-specific \erga. As shown in Table~\ref{tab:cross_platform_transfer}, the transferred \varc retains $95\%$ and $98\%$ of same-configuration mAP on Spheriverse and PanoMMOcc, respectively ($0.1800$ vs.\ $0.1887$ and $0.5803$ vs.\ $0.5897$). This shows that object-level reasoning is preserved after transfer.

\section{Conclusion}
\label{sec:conclusion}
We present \method for perspective-to-ERP adaptation of VFM-based 3D detection. 
The framework addresses the three requirements of this central challenge: \erga aligns pretrained foundation geometry with ERP-specific structure, \varc establishes an explicit hypothesis-evidence-action reasoning chain for evidence-grounded object-level 3D reasoning, and an \aghe preserves local cues for directed heading estimation.
\method achieves the best mAP and NDS on two ERP benchmarks with heterogeneous sensing configurations. \varc also retains strong performance across configurations when paired with target-specific ERGA-Ray. 
These results highlight the effectiveness of \method for mobile embodied agents.
Future work will explore extending evidence-grounded object-level 3D reasoning to temporal panoramic observations and more complex embodied perception tasks.

% The bibliography is emitted automatically once main.bib is added.
\IfFileExists{main.bib}{%
  {\small
  \bibliographystyle{IEEEtran}
  \bibliography{main}
  }%
}{}

\end{document}